\documentclass[10pt,twocolumn,letterpaper]{article}

\usepackage[pagenumbers]{cvpr} 

\usepackage{graphicx}
\usepackage{amsmath}
\usepackage{amssymb}
\usepackage{booktabs}

\usepackage[pagebackref,breaklinks,colorlinks]{hyperref}

\usepackage[capitalize]{cleveref}
\crefname{section}{Sec.}{Secs.}
\Crefname{section}{Section}{Sections}
\Crefname{table}{Table}{Tables}
\crefname{table}{Tab.}{Tabs.}

\usepackage{cuted}
\usepackage{scalerel}

\begin{document}

\title{One-Shot Face Video Re-enactment using \\Hybrid Latent Spaces of StyleGAN2}

\author{Trevine Oorloff and Yaser Yacoob\\
University of Maryland, College Park, MD 20742, USA\\
{\tt\small \{trevine,yaser\}@umd.edu}
}
\maketitle

\begin{strip}\centering
\includegraphics[width=0.8\linewidth]{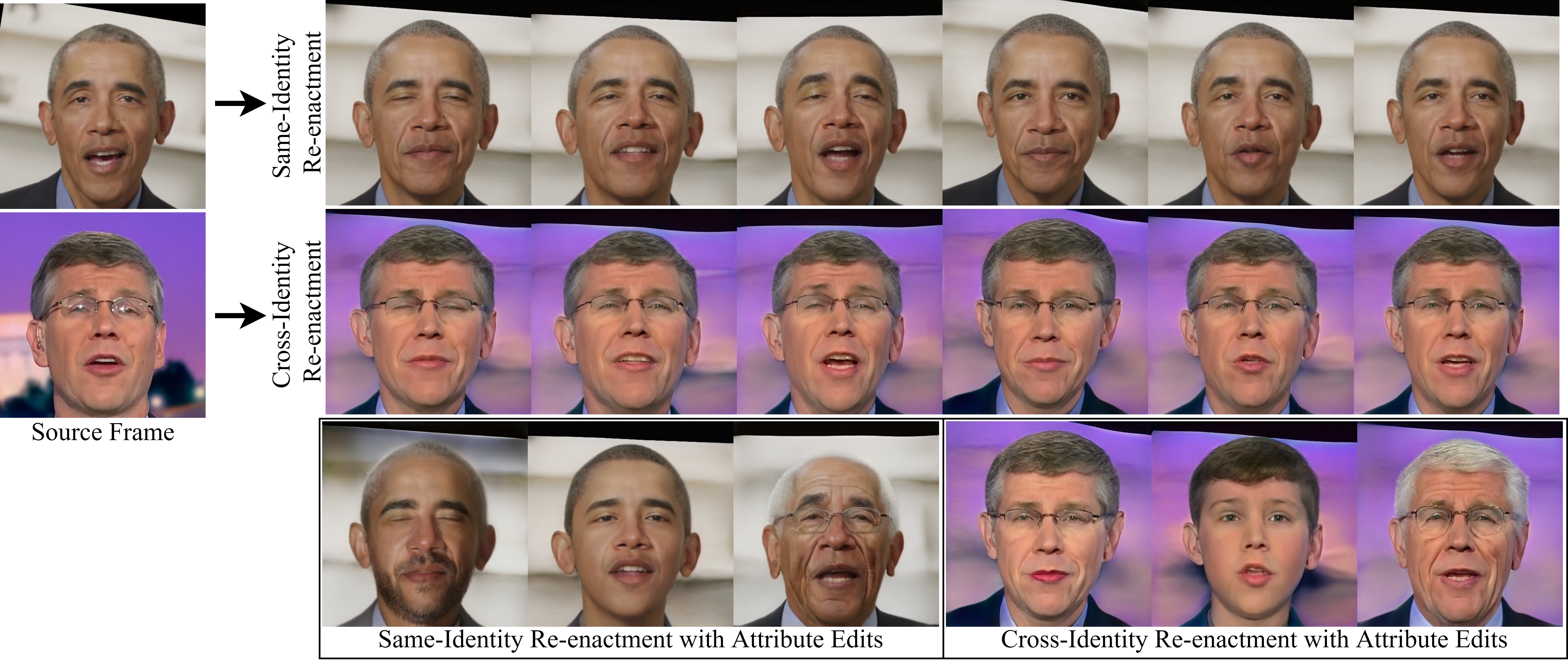}
\captionof{figure}{\label{fig:teaser} \textbf{The proposed end-to-end framework facilitates one-shot face video re-enactment at 1024\textsuperscript{2}, purely based on StyleGAN2's latent spaces without explicit priors.} It also enables temporally consistent latent-based attribute edits such as age, beard, and make-up to be applied on the re-enactments. Observe similar head-pose and expression in re-enacted frames from \textit{Top} to \textit{Bottom} in a column. 
\vspace{-0.05in}
}
\end{strip}

\begin{abstract}

While recent research has progressively overcome the low-resolution constraint of one-shot face video re-enactment with the help of StyleGAN's high-fidelity portrait generation, these approaches rely on at least one of the following: explicit 2D/3D priors, optical flow based warping as motion descriptors, off-the-shelf encoders, \etc, which constrain their performance (\eg, inconsistent predictions, inability to capture fine facial details and accessories, poor generalization, artifacts). 
We propose an end-to-end framework for simultaneously supporting face attribute edits, facial motions
and deformations, and facial identity control for video generation.
It employs a hybrid latent-space that encodes a given frame into a pair of latents: Identity latent, $\mathcal{W}_{ID}$, and Facial deformation latent, $\mathcal{S}_F$, that respectively reside in the $W+$ and $SS$ spaces of StyleGAN2. Thereby, incorporating the impressive editability-distortion trade-off of $W+$ and the high disentanglement properties of $SS$. 
These hybrid latents employ the StyleGAN2 generator to achieve high-fidelity face video re-enactment at $1024^2$. 
Furthermore, the   model supports the generation of realistic re-enactment videos with other latent-based semantic edits (\eg, beard, age, make-up, \etc). 
Qualitative and quantitative analyses performed against state-of-the-art methods demonstrate  the superiority of the proposed approach. The project page is located at \url{https://trevineoorloff.github.io/FaceVideoReenactment_HybridLatents.io/}.
\end{abstract}

\section{Introduction}
\label{sec:intro}

One-shot face video re-enactment refers to the process of generating a video by animating the identity of a portrait image (source frame) mimicking the facial deformations and head pose of a driving video. 
The increasing interest in virtual reality   has stimulated video re-enactment due to its wide range of applications (\eg, digital avatars, animated movies, telepresence). 

The task of one-shot face re-enactment is challenging since it requires extraction of (1) identity and 3D facial structure of the given 2D source frame and (2) motion information from the driving frames, to facilitate realistic animations, despite the unavailability of paired data.
Most common approaches include the use of 2D landmarks \cite{tripathy2021facegan,nirkin2019fsgan,zakharov2020fast, huang2020learning}, 3D parameterized models \cite{yin2022styleheat,kang2022megafr, doukas2021headgan,ren2021pirenderer}, or latents \cite{wang2022latent,burkov2020neural, zeng2020realistic, wiles2018x2face} to capture the underlying facial structure and/or motion. 
Employing strict facial-structure priors may support rigorous control of the facial structure and motion, but these priors suffer from lack of generalizability (for different face geometries), inability to capture fine/complex facial deformations (\eg wrinkles, internal mouth details such as tongue and teeth), inability to handle accessories such as eyeglasses, and inconsistencies in prediction which hinder their performance. 
In addition to latent-based models, research such as \cite{siarohin2019first, wang2021one} proposed models that alleviate the dependency on pre-defined priors by predicting 2D/3D keypoints employing additional networks in an unsupervised manner. 
Even though such models improved generalization, they are limited to producing low resolution videos (mostly $256^2$, but some $512^2$).

StyleGAN's \cite{karras2020analyzing} ability to produce high-resolution ($1024^2$) photo-realistic faces, richness and semantic interpretability of its latent spaces \cite{oorloff2022encode,wu2021stylespace,harkonen2020ganspace, shen2020interface}, and the improvements in inversion techniques  contributed  
to improved re-enactment generations \cite{oorloff2022encode, yin2022styleheat, kang2022megafr}. 
However, both \cite{kang2022megafr} and \cite{yin2022styleheat} use 3D parameterized models to capture the deformations of the facial attributes and thus share the drawbacks of using a pre-defined structural prior as discussed previously. 

Considering the latent space manipulations in \cite{wu2021stylespace, harkonen2020ganspace, oorloff2022encode, abdal2021styleflow} it is evident that the latent space  of a pre-trained StyleGAN has implicit 3D information embedded within it. We conjecture that the StyleGAN's latent spaces are not yet fully exploited  for  re-enactment and the use of explicit structural representations is redundant and limits the performance of StyleGAN to the capacity-limits of such structural priors. While \cite{oorloff2022encode} encodes the facial deformations of the driving sequence directly on the Style-latent space, it follows an optimization-based approach to deduce the deformation encoding which is time-consuming similar to other optimization-based approaches which limits its practicality. 

We  address the following question: \textit{Can we learn a general  model to facilitate face identity, attributes, and motion edits exploiting the latent spaces of StyleGAN2 without reliance on explicit 2D/3D facial structure models while improving the performance of generating realistic, high-quality, and temporally consistent one-shot face videos?} 

We propose a novel framework that encodes a portrait image as an Identity latent, $\mathcal{W}_{ID}$, and a Facial deformation latent, $\mathcal{S}_F$, that reside in the pre-defined latent spaces of StyleGAN2. This encoding  not only facilitates high-quality high-resolution ($1024^2$) one-shot face re-enactment (both same and cross-identity) through the StyleGAN2 generator, but is also capable of generating re-enactment videos with realistic facial edits (\eg, beard, age, make-up) accommodating latent manipulation techniques such as \cite{shen2020interface, harkonen2020ganspace}. 

Considering the three prominent intermediate latent spaces of StyleGAN2: $W$, $W+$, and $SS$: the $W$ space suffers from poor inversion; the $W+$ space has the best trade-off between inversion quality and editability as demonstrated by StyleGAN inversion and latent manipulation research \cite{tov2021designing, alaluf2021restyle, richardson2021encoding}; and the StyleSpace, $SS$, is the most disentangled \cite{wu2021stylespace}. We combine the $W+$ and $SS$ spaces into a hybrid space, so that the Identity latent, $\mathcal{W}_{ID}$, and face feature deformation latent, $\mathcal{S}_F$, are learned by encoders that capture the richness embedded in these latent spaces. Thereby, simultaneously supporting face attribute edits, facial motions and deformations, and facial identity control. 
In summary, our key contributions include:
\begin{itemize}
    \vspace{-0.1in} 
     \setlength{\itemsep}{0pt}
    \setlength{\parskip}{2pt}
    \setlength{\parsep}{2pt}
    \item A novel framework that enables high-fidelity robust one-shot face re-enactment (same and cross-identity) video generation at $1024^2$, which also facilitates realistic facial edits such as age, beard, and make-up, 
    \item A novel approach of using a combination of two pre-defined latent spaces ($W+$ and $SS$) of StyleGAN2 to have zero dependencies on explicit structural facial priors such as 2D/ 3D landmarks or parameterizations,
    \item A novel   ``Cyclic Manifold Adjustment" (CMA), that locally adjusts the StyleGAN2's manifold to improve the reconstruction of an out-of-domain source and enable seamless transfer of facial deformations of the driving video to the source image. 
\end{itemize}

\section{Related Work}
\label{sec:related_work}

\textbf{Latent Space Manipulation of StyleGAN:} Since the proposal of StyleGAN, there has been a plethora of research on the semantic interpretability of the intermediate latent spaces \cite{shen2020interface,harkonen2020ganspace,wu2021stylespace, shen2020interpreting}. Improvements in GAN inversion techniques \cite{abdal2019image2stylegan, tov2021designing, alaluf2021restyle, roich2021pivotal, wang2022hfgi}  complemented such research facilitating realistic edits of real-world images through latent-space manipulations. 

Semantic editing using latent spaces has employed supervised \cite{shen2020interface, wu2021stylespace, abdal2021styleflow}  and unsupervised approaches \cite{harkonen2020ganspace, voynov2020unsupervised, peebles2020hessian}. These methods enable realistic facial edits such as head-pose, expression, gaze, gender, age, and eyeglasses by traversing the latent space of a pre-trained StyleGAN. 

\begin{figure*}[t!]
    \centering
    \includegraphics[width=0.75\linewidth]{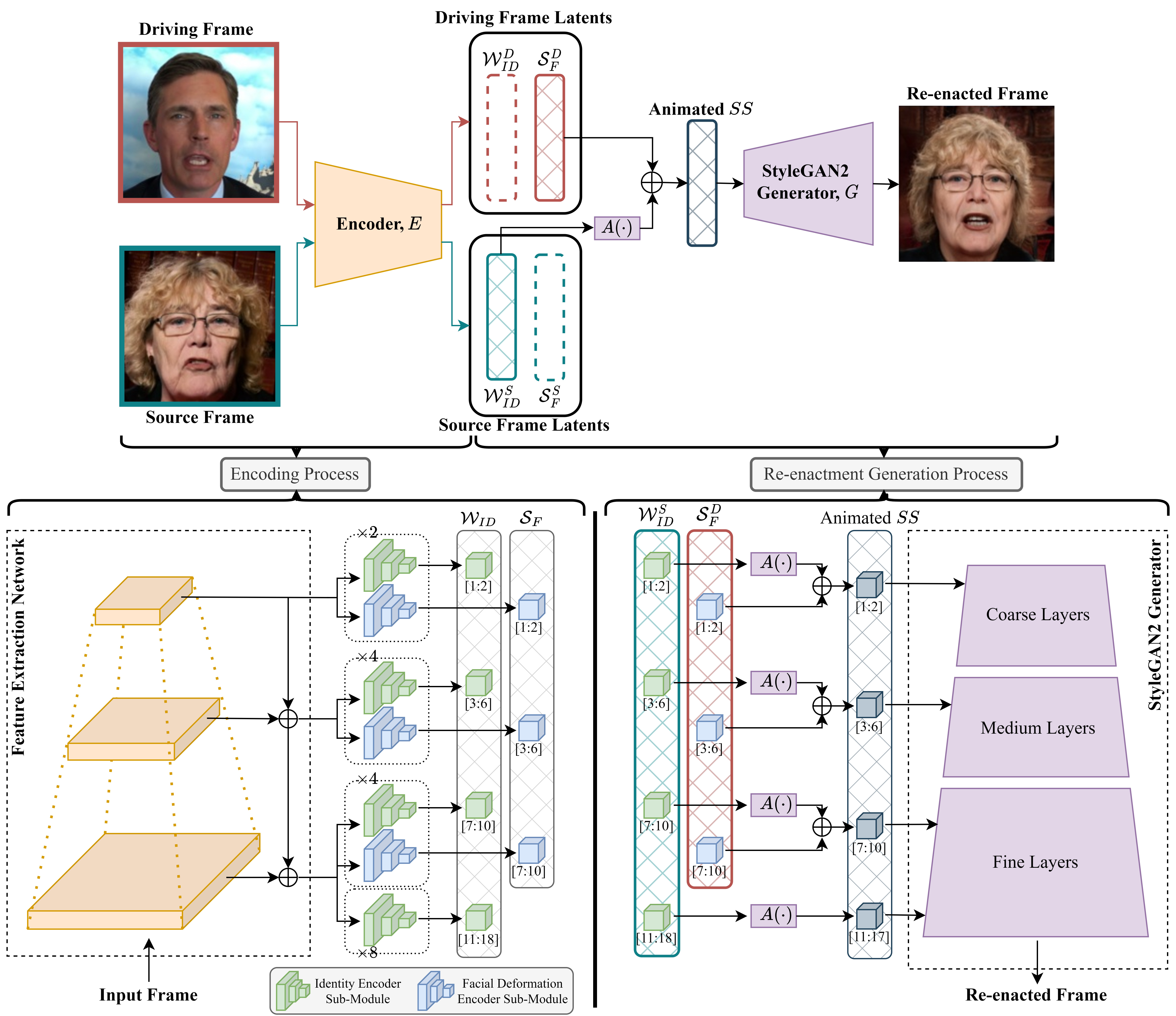}
    \caption{\textbf{The pipeline of the proposed framework.} The high-level re-enactment process (\textit{Top}), the expanded architectures of the encoding (\textit{Bottom-Left}) and re-enactment (\textit{Bottom-Right}) processes are depicted. In encoding, given a frame, the Encoder, $E$, outputs a pair of latents: Identity latent, $\mathcal{W}_{ID}$, and Facial-deformation latent, $\mathcal{S}_F$. In re-enactment, $\mathcal{S}_F^D$ (driving frame) is added to $\mathcal{W}_{ID}^S$ (source),  transformed using $A(\cdot)$ to obtain the animated $SS$ latent, which is used to obtain the re-enacted frame using the StyleGAN2 Generator, $G$. }
    \label{fig:pipeline}
    \vspace{-0.1in}
\end{figure*}

\textbf{Face Video Re-enactment:}
Face video re-enactment approaches can be categorized based on the identity/motion representation or the approach used to generate the animated frames. The common approaches used for identity/motion representation include facial landmarks \cite{tripathy2021facegan,nirkin2019fsgan,zakharov2020fast, huang2020learning}, 3D facial priors \cite{kang2022megafr, doukas2021headgan,ren2021pirenderer}, 2D/3D predictive keypoint methods \cite{siarohin2019first,wang2021one}, and intermediate latent representations \cite{wang2022latent,burkov2020neural, zeng2020realistic, wiles2018x2face}. 
While 3D facial priors address the main issue of 2D facial priors, \ie, unrealistic deformations caused during significant motion, their performance is limited by the lack of fine visual-details surrounding face dynamics (wrinkles, tongue and teeth, non-rigid deformations), inability to represent accessories such as eyeglasses, and representation of only the inner face region.
Further, the  use of 2D/3D facial priors leads to spatio-temporal incoherence stemming from the inconsistencies of the landmarks/parameters and poor generalization in cases of varying facial geometries between the driving and source identities. Even though keypoint predictive networks and latent-based methods have improved generalization, they have been limited to low-resolution video generation. 

The work on re-enactment either follows an optical flow based warping strategy \cite{siarohin2019first, wang2021one, doukas2022free} or a generative approach \cite{wang2022latent, hsu2022dual, burkov2020neural} to   animate  frames. Even though warping methods yield results with high resemblance to the in-the-wild source images due to operating in image space, they could cause unrealistic deformations in faces, have weaker generalization compared to generative approaches, and perform poorly in generating facial structures that were not visible in the source image (\eg, filling in teeth in opening of the mouth, opening eyes, ears when rotating the head).

In contrast to previous approaches, LIA \cite{wang2022latent} utilizes a latent-space navigation based generative approach to facilitate re-enactment. However, the  results are limited to $256^2$ and require  the pose of the source and the initial driving frame to be of similar pose which limits the practicality.

\textbf{StyleGAN-based Face Re-enactment:}
Recent research \cite{kang2022megafr, yin2022styleheat, bounareli2022finding} employed StyleGAN2 for high-resolution one-shot face video re-enactment due to its ability to produce realistic portraits at $1024^2$ and the rich semantic properties of its latent spaces.  
MegaFR \cite{kang2022megafr}   encodes the residual deformation between the source image and a 3D rendering of the animated frame as an additive $W+$ space offset. The rendering of the animated frame is a parameterized approximation obtained using a combination of 3DMM parameters of the source and driving frames. Bounareli \etal in \cite{bounareli2022finding} map the difference between the output parameters of a 3D model onto the $W+$ space to obtain the re-enacted frame's latent. StyleHEAT \cite{yin2022styleheat} uses 3DMM parameters to capture the facial deformations to generate flow fields which are used to warp an intermediate feature space of the generator. 

While the above methods yield promising results, they employ 3D models to parameterize the motion, thus, suffer from the limitations of using 3D priors as explained previously (\ie, inconsistencies, lack of fine-grained details, limited by the capacity of the 3D model, \etc). 
Our proposed model does not use \textit{explicit} 2D/ 3D structural models, instead, we exploit the  editability and disentanglement of the latent spaces ($W+$ and $SS$) and its  \textit{implicit} 3D priors to encode both the identity and the motion within the StyleGAN2's pre-defined latent spaces. Importantly, we propose a unified end-to-end system  trained using a self-supervised approach and hence is not bounded by the limitations of other models (\eg, \cite{yin2022styleheat} and \cite{kang2022megafr} rely on inversion models to obtain the source latent, \cite{yin2022styleheat,kang2022megafr, bounareli2022finding} rely on explicit 3D models). While StyleGAN2 generation inherently suffers from texture-sticking, StyleHEAT aggravates the issue as their model warps the intermediate feature space of the StyleGAN2 generator.  
Moreover, compared to \cite{kang2022megafr, bounareli2022finding} that encode  in the $W+$ space, we employ a hybrid latent space approach using both $W+$ and $SS$ latent spaces to exploit high inversion-editability and high disentanglement respectively.

\section{Methodology}
\label{sec:method}

\textbf{Preliminaries:}
The StyleGAN2 generator  consists of several latent spaces: $Z$, $W$, $W+$, and $SS$. The vanilla generation process consists of an initial random noise latent, $z \in Z \sim \mathcal{N}(\mu,\,\sigma^{2})$,
which is mapped to the $w \in W$ latent using a sequence of fully connected networks, which is then transformed using layer-wise affine transformations, $A(\cdot)$, to form the StyleSpace, $SS$. While this model uses the same $w \in W$ latent to each of the transformations in $A(\cdot)$, StyleGAN2 inversion methods such as \cite{richardson2021encoding, tov2021designing,alaluf2021restyle} use an extension of the $W$ space defined as $W+$, where the input to each transformation in $A(\cdot)$ is allowed to differ in order to improve the reconstruction quality. While there are multiple approaches for forming $W+$, we refer to the $W+$ space in \cite{richardson2021encoding} due to its impressive balance between the distortion-editability trade-off. Further, as evaluated by \cite{wu2021stylespace}, the $SS$ has the best disentanglement, completeness, and informativeness scores among $W$, $W+$, and $SS$ spaces thus making it the best latent-space for disentangled manipulations.

\textbf{Overview:}
As shown in \cref{fig:pipeline}, the proposed framework comprises of two main networks, an encoder, $E$, and a decoder, $G$: a pre-trained StyleGAN2 generator. The encoder consists of two heads: Identity Encoder, $E_{ID}$, and Facial Deformation Encoder, $E_{F}$, preceded by a common Feature Extraction Network, $F$. Given an input frame, the encoder creates two latents, $\mathcal{W}_{ID}$ and $\mathcal{S}_F$, capturing identity and facial deformations respectively, such that,
\begin{gather}
    \mathcal{W}_{ID} = E_{ID}(F(I)), \quad \mathcal{S}_F = E_F(F(I)), \label{eq:encode} \displaybreak[0]\\
    I = G \left\{ A(\mathcal{W}_{ID}) + \mathcal{S}_F \right\}. \label{eq:img_gen}
\end{gather}
While  $\mathcal{W}_{ID}$  resides in the entire $W+$ space \ie, $\mathcal{W}_{ID} \in W+ \in \mathbb{R}^{18\times512}$,  $\mathcal{S}_{F}$ resides in the space spanned by the first 10 $SS$ layers \ie, $\mathcal{S}_{F} \in SS_{1:10} \in \mathbb{R}^{10\times512}$. This is based on (1) the observation in \cite{oorloff2022encode, wu2021stylespace} that $SS$ latents corresponding to facial deformations of interest, (\ie, pose, mouth, gaze, eyes, eyebrows, and chin) lie within the first $10$ layers of $SS$ and (2) avoiding the appearance jitters caused by edits on high-resolution feature layers \cite{yao2022feature}. 

In re-enactment, we follow a frame-wise approach, where a single source frame, $I^S$, and each frame, $I^D_t$, of the driving sequence are projected to $\{\mathcal{W}_{ID}^S, \mathcal{S}_{F}^S\}$ and $\{\mathcal{W}_{ID_t}^D, \mathcal{S}_{F_t}^D\}$ respectively (using \cref{eq:encode}). Thereafter, the animated frame, $I^{S\rightarrow D}_t$ is generated using $G$, sourcing $\mathcal{W}_{ID}^S$ and $\mathcal{S}_{F}^D$ latents, comprising of the source identity and the driving frame's facial deformations respectively.
\begin{equation}\label{eq:reenact}
    I^{S\rightarrow D}_t = G\left\{ A(\mathcal{W}_{ID}^S) + \mathcal{S}_{F_t}^D\right\}
\end{equation}

As seen in \cref{eq:reenact}, the additive latent $\mathcal{S}_{F_t}^D$ constitutes a latent edit performed on  $\mathcal{W}_{ID}^S$. 
Thus, it is important for $\mathcal{W}_{ID}^S$ to reside in the well-behaved editable regions of the latent spaces of StyleGAN2 (\ie, $W+$) and accommodate a wide range of face deformation latent edits imposed by the driving sequence ($\{\mathcal{S}_{F_t}^D\}$) and for  $\mathcal{S}_{F_t}^D$ to reside in a highly disentangled region (\ie $SS$) of the latent spaces, so that it minimizes identity leakage and altering the source identity across frames. This design enables the manipulation of attributes such as age, beard, make-up, \etc. through latent space edits proposed by \cite{shen2020interface,harkonen2020ganspace}. 
\begin{equation}\label{eq:edit}
    I^{S\rightarrow D}_{edit} = G \left\{ A(\mathcal{W}_{ID}^S + \mathcal{W}_{edit}) + \mathcal{S}_{F}^D \right\}
\end{equation}

\subsection{Architecture}
\textbf{Feature Extraction Network, $F$:}
We use a ResNet50-SE backbone \cite{he2016resnet,hu2018squeeze} extended with a feature pyramid \cite{lin2017feature} to extract the coarse, medium, and fine features of each frame similar to \cite{richardson2021encoding}. These levels  correspond to the levels of features addressed by each latent layer as  in \cite{karras2019style}. 

\textbf{Identity Encoder, $E_{ID}$:} 
The $E_{ID}$ consists of network blocks similar to ``map2style" in \cite{richardson2021encoding} where the feature maps of the corresponding granularity are gradually reduced to $\mathbb{R}^{512}$ using a fully convolutional network. The encoder consists of $18$ such blocks each predicting a single layer (dimension) of $\mathcal{W}_{ID} \in \mathbb{R}^{18\times512}$.

\textbf{Facial Deformation Encoder, $E_{F}$:}
While $E_{F}$ has a similar architecture to $E_{ID}$, it consists of only $10$ ``map2style" blocks as we limit the $SS$ latent edits to only the first 10 layers of $SS$ as explained above. 

\textbf{Decoder, $G$:}
We use the pre-trained StyleGAN2 generator, which facilitates the input of $SS$ latents \cite{wu2021stylespace}, as the decoder to generate the re-enacted frames from the latents.

\subsection{Implementation}

Due to the unavailability of paired re-enactment datasets, we follow a self-supervised training approach to learn the weights of the encoder, $E$. During training, we randomly sample a single source frame, $I^S$, and two driving frames, $I^{D1}$ and $I^{D2}$, one belonging to the same identity as $I^S$ and the other from a randomly selected different identity respectively. 
The three frames, $I^S$, $I^{D1}$, and $I^{D2}$ are sent through the encoder, $E$, to obtain the corresponding latents, $\{ \mathcal{W}^S_{ID}, \mathcal{S}_F^S\}$, $\{ \mathcal{W}^{D1}_{ID}, \mathcal{S}_F^{D1}\}$, and $\{ \mathcal{W}^{D2}_{ID}, \mathcal{S}_F^{D2}\}$ respectively (using \cref{eq:encode}). We learn the weights of $E$ by optimizing over the following loss functions: 

\textbf{Reconstruction Losses:} 
The reconstruction losses are two-fold comprising of a self-reconstruction loss, \cref{eq:l_self}, and a re-enactment loss, \cref{eq:l_reenact}, which measure the reconstruction of the source frame, $I^{S\rightarrow S}$, and the same-identity driving frame, $I^{S \rightarrow D1}$, using the source identity latent, $\mathcal{W}_{ID}^S$, and the corresponding facial deformation latents.
\begin{gather}
    \mathcal{L}_{self} = \mathcal{L}_{rec} \left\{ I^{S} , G \{ A(\mathcal{W}_{ID}^S) + \mathcal{S}_F^{S} \}\right \} \label{eq:l_self}  \displaybreak[0]\\
    \mathcal{L}_{reenact} = \mathcal{L}_{rec} \left\{ I^{D1} , G \{ A(\mathcal{W}_{ID}^S) + \mathcal{S}_F^{D1} \}\right \} \label{eq:l_reenact}  \displaybreak[0]\\
    \mathcal{L}_{rec} = \lambda_{L2} \mathcal{L}_{L2}  + \lambda_{LPIPS} \mathcal{L}_{LPIPS} + \lambda_{GV} \mathcal{L}_{GV}
\end{gather}
where, $\mathcal{L}_{rec}$ is a weighted sum of the MSE loss, $\mathcal{L}_{L2}$, LPIPS loss \cite{zhang2018perceptual}, $\mathcal{L}_{LPIPS}$, and Gradient Variance loss \cite{abrahamyan2022gradient}, $\mathcal{L}_{GV}$, weighed by $\lambda_{L2}$, $\lambda_{LPIPS}$, and $\lambda_{GV}$ respectively.

\textbf{Identity Loss:} 
The identity loss is computed using, 
\begin{align}
    \mathcal{L}_{id} &= \{1 - \left \langle  \phi(I^S) , \phi(I^{S_{ID}}) \right \rangle  \} \label{eq:l_id_s} \\
                    &+ \{1 - \left \langle  \phi(I^S) , \phi(I^{S \rightarrow D2}) \right \rangle \} \label{eq:l_id_p2p}
\end{align}
where, the cosine similarity ($\langle \cdot , \cdot \rangle $) of the ArcFace\cite{deng2019arcface} feature space is measured between the pair of images. $I^{S_{ID}} = G \{ A(\mathcal{W}_{ID}^S) \}$ is the source identity image and $I^{S \rightarrow D2} = G \{ A(\mathcal{W}_{ID}^S) + \mathcal{S}_F^{D2}\}$ denotes the re-enactment of the source representing the facial deformations of $I^{D2}$. \cref{eq:l_id_s} ensures the identity of the source is captured within $\mathcal{W}_{ID}^S$ and also prevents the optimization from converging to the trivial solution of \cref{eq:l_self,eq:l_reenact}, which is  $\mathcal{W}_{ID}^S = 0$. Further, \cref{eq:l_id_p2p} makes sure that puppeteering preserves identity \ie, minimizes the identity information leakage to $\mathcal{S}_F$.

\textbf{Identity Latent Consistency Loss:}
For additional control over puppeteering, to encourage consistent identity latents irrespective of head-pose and facial attributes of the source, we obtain the $\mathcal{W}_{ID}$ of $I^{S\rightarrow D2}$ by passing it through $E$ and compute the following loss over the latent space.  
\begin{equation}\label{eq:l_wid}
    \mathcal{L}_{w\_id} = \lVert {W}_{ID}^{S} - {W}_{ID}^{S \rightarrow D2} \rVert_2
\end{equation}

\textbf{Regularization Loss:}
Additional regulatory losses are used to reduce the variance within the $\mathcal{W}_{ID}^S$ \cite{tov2021designing} and to control the facial-deformation edits, $\mathcal{S}_F^{D1}$, to be within the proximity of $A(\mathcal{W}_{ID}^S)$ combined in a ratio of $1:\lambda _{S}$.  $\mathcal{W}_{ID}^S [i]$ and $\mathcal{S}_{F}^{D1} [j]$ correspond to the $i^{th}$ and $j^{th}$ dimension of $\mathcal{W}_{ID}^S$ and $\mathcal{S}_{F}^{D1}$ latents respectively.
\begin{gather}
    \Delta_i = \mathcal{W}_{ID}^S [i] - \mathcal{W}_{ID}^S [1] \displaybreak[0]\\
    \mathcal{L}_{reg} = \sum_{i=1}^{18} \{ \lVert \Delta_i \rVert _2 \} 
    + \lambda_{S}\sum_{j=1}^{10} \{ \lVert \mathcal{S}_{F}^{D1} [j] \rVert _2 \}
\end{gather}

\textbf{Feature Reconstruction Loss:} 
Complementary to the reconstruction losses, feature reconstruction losses are computed using the same loss functions with the exception of the losses being computed on a dilated masked region consisting of the mouth, eyes, and eyebrows to increase the emphasis on capturing facial deformations accurately.
\begin{align}
    \mathcal{L}_f &= \mathcal{L}_{rec} \left\{ M^S \odot I^{S} , M^S \odot I^{S\rightarrow S}\right \} \nonumber  \\
    &+ \mathcal{L}_{rec} \left\{ M^{D1} \odot I^{D1} , M^{D1} \odot I^{S\rightarrow D1} \right \} \label{eq:l_features}
\end{align}

Additionally, similar to \cite{tov2021designing} we train a \textbf{Latent Discriminator}, with an adversarial loss, $\mathcal{L}_d$, to encourage the $W_{ID}^S$ latents to be in the well-editable regions of the StyleGAN2 latent space.  

\textbf{Total Loss:} 
The total loss is as follows, where $\lambda_*$ represents the corresponding weights.
\begin{gather}
    \mathcal{L} = \mathcal{L}_{self} + \mathcal{L}_{reenact} + \lambda_{id}\cdot\mathcal{L}_{id} + \lambda_{w\_id}\cdot\mathcal{L}_{w\_id} \nonumber \\
    + \lambda_{d}\cdot\mathcal{L}_{d} + \lambda_{reg}\cdot\mathcal{L}_{reg} + \lambda_{f}\cdot\mathcal{L}_{f}
\end{gather}

\begin{figure}[b!]
    \vspace{-0.1in}
    \centering
    \includegraphics[width=0.7\linewidth]{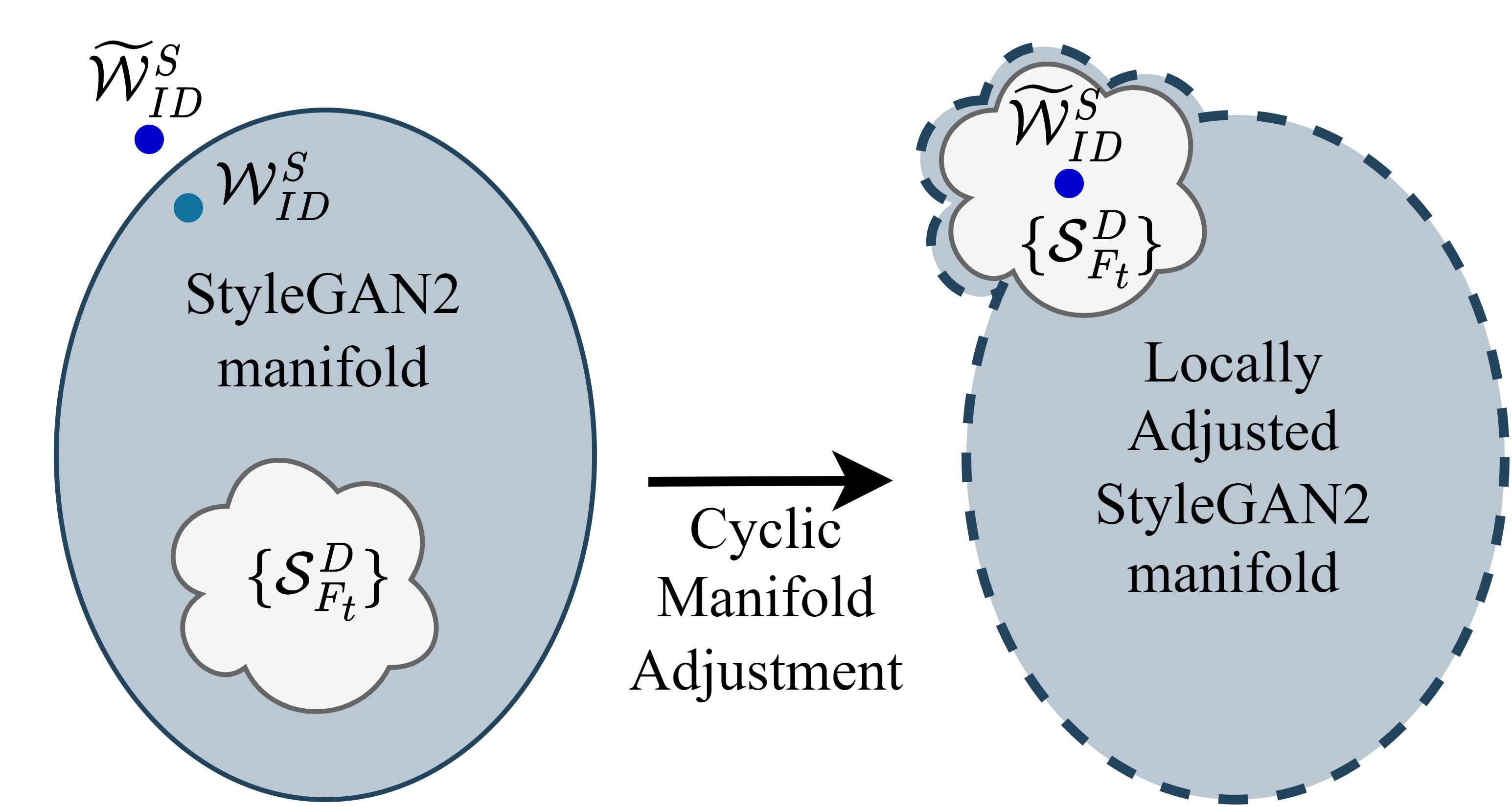}
    \caption{ \textbf{Cyclic Manifold Adjustment (CMA).} For an out-of-domain subject, $\widetilde{\mathcal{W}}_{ID}^S$, $\mathcal{W}_{ID}^S$, and $\{\mathcal{S}_{F_t}\}$, represent the true identity, identity latent estimate obtained using  $E$, and the sequence of facial deformation latents obtained from the driving sequence. We locally tweak the StyleGAN2's manifold around $\mathcal{W}_{ID}^S$, to include the latent space spanned by $\{\mathcal{S}_{F_t}\}$ centered around $\widetilde{\mathcal{W}}_{ID}^S$, thus improving the source identity reconstruction and enabling seamless transfer of facial deformations of the driving video. 
    }
    \label{fig:cyc_manifold_adj }
    \vspace{-0.1in}
\end{figure}

\begin{figure*}[t!]
  \centering
  \includegraphics[width=0.95\textwidth]{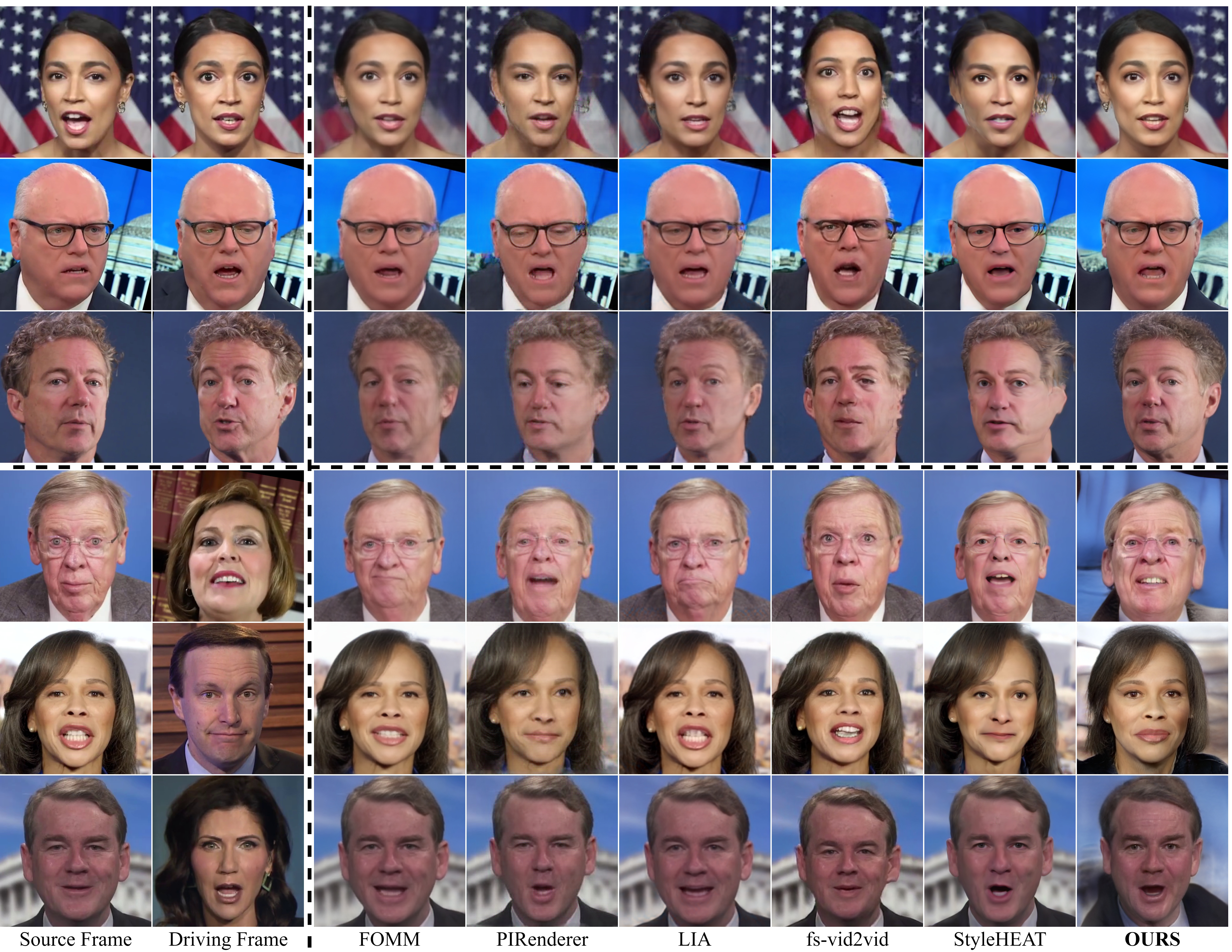}
  \caption{\label{fig:reenact} \textbf{Qualitative evaluation of same-identity \textit{(Top)} and cross-identity \textit{(Bottom)} re-enactment}. \underline{Same-Identity Re-enactment:} Observe the lack of sharpness in facial features (\eg teeth, wrinkles, eyes), visual artifacts around eyes, ears, and mouth, and incorrect facial features in baseline methods in comparison to our approach. \underline{Cross-Identity Re-enactment:}  Observe in comparison to the baselines: \textit{4th row:} teeth, mouth formation, and head-pose; \textit{5/6th row:} preservation of source identity and lip structure, and the expression of driving.
  }
  \vspace{-0.1in}
\end{figure*}

\subsection{Cyclic Manifold Adjustment (CMA)}\label{sub:cyc_manifold}

StyleGAN2 based approaches have a comparatively weaker identity reconstruction for out-of-domain subjects. While several methods \cite{wang2022hfgi, roich2021pivotal} have been proposed to improve the reconstruction quality of real-world images, none of them could be directly used for cross-identity re-enactment due to the unavailability of paired-data. The use of such approaches on the source image alone results in subpar performance and/or generates visual artifacts when the facial deformation latents ($\{\mathcal{S}_{F_t}^D\}$) are added.

To improve the identity reconstruction quality of out-of-domain subjects in a cross-identity re-enactment setting, we propose a novel approach, ``Cyclic Manifold Adjustment", inspired by PTI \cite{roich2021pivotal}. Suppose the true source identity latent is $\widetilde{\mathcal{W}}_{ID}^S$ which is out of StyleGAN2's domain.
Fine-tuning the StyleGAN2 generator, $G$, using PTI on the source image, would constrain the local latent space around the source identity, $\widetilde{\mathcal{W}}_{ID}^S$, thus limiting the editability through facial deformation latents of the driving frames, $\{\mathcal{S}_{F_t}^D\}$. 
In contrast, Cyclic Manifold Adjustment tweaks the latent space manifold around the source identity latent estimate, $\mathcal{W}_{ID}^S$ obtained through $E$, to include the latent space spanned by the sequence $\{\mathcal{S}_{F_t}^D\}$ centered around $\widetilde{\mathcal{W}}_{ID}^S$ as depicted in \cref{fig:cyc_manifold_adj }. 
This novel approach  improves the identity reconstruction of the out-of-domain source and enables seamless transfer of facial deformations of the driving video.
We achieve this by optimizing the cost function,
\begin{equation}
    \mathcal{L} \{ I^S_{cyc}, I^S  \} + \mathcal{L} \{ I^D_{cyc}, I^D  \}
\end{equation}
where, $\mathcal{L}$ denotes a combination of LPIPS and L2 losses and $I^S_{cyc}$ and $I^D_{cyc}$ are cyclic reconstructions of the source and driving frames respectively generated as follows.
\begin{gather}
    \mathcal{W}_{ID\_{cyc}}^S, \, \mathcal{S}_{F\_{cyc}}^D = E ( I^{S\rightarrow D} ) \label{eq:cyc_manifold_1}  \displaybreak[0]\\
    I^S_{cyc} = G\left\{ A(\mathcal{W}_{ID\_{cyc}}^S) + \mathcal{S}_{F}^S\right\} \label{eq:cyc_manifold_2}  \displaybreak[0]\\
    I^D_{cyc} = G\left\{ A(\mathcal{W}_{ID}^D) + \mathcal{S}_{F\_{cyc}}^D\right \}  \label{eq:cyc_manifold_3}
\end{gather}

\begin{table*}[t!]
    \centering
    \resizebox{0.95\linewidth}{!}{ 
    \begin{tabular}{ l  | c c c c c c c c c c c c } 
         \hline
         Method & res. & L1 $\downarrow$ & LPIPS$\downarrow$ & $\mathcal{L}_{ID}$$\downarrow$ & PSNR$\uparrow$  & SSIM$\uparrow$ &FID$\downarrow$ & FVD$\downarrow$  & FVD\textsubscript{M}$\downarrow$ & $\rho_{\scaleto{AU}{3pt}}$$\uparrow$ &
         $\rho_{\scaleto{GZ}{3pt}}$$\uparrow$ &
         $\rho_{\scaleto{pose}{3pt}}$$\uparrow$\\  
         \hline
         FOMM \cite{siarohin2019first}& $256^2$ & \underline{2.37} & \underline{0.042} & \textbf{0.095} & \underline{32.8} & \underline{0.959} &  \underline{22.3} & \underline{102.1} & 19.9 & 0.808 & 0.737 & \underline{0.948}\\
         PIRenderer \cite{ren2021pirenderer} & $256^2$ & 3.52 & 0.053 & 0.118 & 29.1 & 0.932 & 28.0 & 145.1 & 26.8 & 0.748 & 0.770 & 0.906 \\
         LIA \cite{wang2022latent}& $256^2$ & 2.72 & 0.049 & 0.102 & 31.5 & 0.951 &  26.4 & 105.0 & \underline{19.6} & \underline{0.822} & 0.732 & 0.944 \\
         fs-vid2vid \cite{wang2019few} & $512^2$ & 4.38 & 0.065 & 0.151 & 27.4 & 0.919 & 31.6 & 255.0 & 45.0 & 0.572 & 0.635 & 0.822 \\
         StyleHEAT \cite{yin2022styleheat}& \textbf{1024\textsuperscript{2}} & 3.59 & 0.059 & 0.133 & 28.8 & 0.934 & 38.8 & 171.0 & 31.4 & 0.745 & \underline{0.850} & 0.921\\
         \textbf{Ours} & \textbf{1024\textsuperscript{2}} & \textbf{2.28} & \textbf{0.027} & \underline{0.097} & \textbf{33.1} & \textbf{0.967} & \textbf{19.3}  & \textbf{101.0} & \textbf{16.6} & \textbf{0.829} & \textbf{0.872} & \textbf{0.952}\\
         \hline
         \textbf{Ours} w/o ID reg & $1024^2$ & 2.34 & 0.028 & 0.111 & 32.8 & 0.953 & 20.3 & 115.4 & 20.7 & 0.810 & 0.857 & 0.948 \\
         \textbf{Ours} w/o Hybrid & $1024^2$ & 2.52 & 0.031 & 0.114 & 31.7 & 0.950 & 20.6& 116.9& 23.6& 0.746 & 0.798 & 0.923\\
         \hline
    \end{tabular}
    }
    \caption{\textbf{Quantitative comparison of one-shot same-identity re-enactment against baselines.} 
    \textit{Top}: Evaluation results computed over 75 unseen videos (37.5K total frames) of the HDTF dataset \cite{zhang2021hdtf}. Our approach yields the best performance across all metrics except $\mathcal{L}_{ID}$ (comparable with best) while generating high-resolution re-enactment. \textit{Bottom}: Ablations performed for \underline{Ours w/o ID reg.:} Framework without the Identity regularization and \underline{Ours w/o Hybrid:} Framework without the Hybrid latent spaces \ie, both latents in $W+$ space.}  
    \label{tab:same_id_reenact}
    \vspace{-0.1in}
\end{table*}

\begin{table}[t]
    \centering
    \resizebox{\linewidth}{!}{
    \begin{tabular}{ l  | c c c c c  }
        \hline 
        Method &FID$\downarrow$& FVD$\downarrow$ & $\text{FVD}_{\scaleto{M}{3pt}}$$\downarrow$ 
        & $\rho_{\scaleto{AU+GZ}{3pt}}$$\uparrow$ & $\rho_{\scaleto{pose}{3pt}}$$\uparrow$\\
        \hline 
        FOMM \cite{siarohin2019first} & 94.0 & 529.4 & 78.4 & 0.450 & 0.782 \\
        PIRenderer \cite{ren2021pirenderer} & \underline{84.8} & 417.3 &  \underline{54.2} & \underline{0.668} & \underline{0.880} \\
        LIA \cite{wang2022latent} & 94.8 & 536.2 & 76.7  & 0.404 & 0.788\\
        fs-vid2vid \cite{wang2019few}  & 90.6 & 532.7 & 86.7  & 0.493 & 0.745 \\
        StyleHEAT \cite{yin2022styleheat} & 97.2 & \underline{408.8} & 58.9  & 0.645 & 0.875 \\
        \textbf{Ours}  & \textbf{74.3} & \textbf{375.4} & \textbf{50.2} & \textbf{0.718} & \textbf{0.915}\\
        \hline
        \textbf{Ours} w/o ID reg  & 85.6 & 399.8 &  53.1 &  0.685 & 0.894\\
        \textbf{Ours} w/o Hybrid  & 93.0 & 427.3 & 60.8 & 0.649 & 0.878 \\
        \textbf{Ours} w/o CMA & 86.9 & 388.1 & 52.4 & 0.646 & 0.896\\
        \textbf{Ours} -- CMA + PTI & 84.8 & 421.2 & 56.8  & 0.627 & 0.890 \\
        \hline
    \end{tabular}
    }
    \caption{\textbf{Quantitative  evaluation of cross-identity one-shot re-enactment.} \textit{Top}: Evaluation computed over 75 unseen videos (37.5K frames in total) of the HDTF dataset \cite{zhang2021hdtf} and random source frames of different identities. Our approach yields the best performance across all metrics which is reflective of the visual results. \textit{Bottom}: Ablations  for \underline{Ours w/o ID reg.:} Framework without the Identity regularization; \underline{Ours w/o Hybrid:} Framework without the Hybrid latent spaces \ie, both latents in $W+$ space; \underline{Ours w/o CMA:} Framework without Cyclic Manfold Adjustment (CMA) and \underline{Ours -- CMA + PTI:} Framework replacing Cyclic Manfold Adjustment (CMA) with PTI\cite{roich2021pivotal}.}
    \label{tab:cross_id_reenact}
    \vspace{-0.1in}
\end{table}

\section{Experiments and Results}\label{sec:experiments}

\textbf{Datasets:} We pre-trained the encoder on the CelebV-HQ dataset \cite{zhu2022celebvhq}, which includes diverse high resolutions (min. $512^2$)  with over 35K videos involving 15K+ identities. The HDTF dataset \cite{zhang2021hdtf}, which consists of 362 videos (720p/1080p) with 300+ identities, was used for the fine-tuning stage with an 80-20 non-overlapping train-test split. 
All the sampled frames were preprocessed according to \cite{siarohin2019first}. For training 50 samples/video were used and for evaluation the first 500 samples/video of 75 unseen videos (total 37.5K frames) were chosen.  

Training was performed in two stages consisting of a pre-training stage followed by a fine-tuning stage. While the former focuses mainly on learning the features of images, improving generalization, and learning the implicit prior of the StyleGAN2's latent space, the latter stage focuses on capturing the detailed facial deformations and face details.

\begin{table}[t!]
    \centering
    \resizebox{0.96\linewidth}{!}{
    \begin{tabular}{ l  | c c c c}
        \hline 
        Method & LPIPS$\downarrow$& $\mathcal{L}_{ID}\downarrow$& FID$\downarrow$& FVD$\downarrow$ \\
        & \small$\times10^{-2}$ & \small$\times10^{-1}$ & \small$\times10^1$ & \small$\times10^2$\\
        \hline 
        FOMM  & $7.5\pm2.1$ & $2.3\pm1.1$ &$3.2\pm1.0$ & $2.0\pm0.7$\\
        PIRenderer  & $\underline{5.7\pm0.4}$ & $\underline{1.2\pm0.2}$ & $\underline{2.2\pm0.2}$ & $1.5\pm0.3$ \\
        LIA  & $7.8\pm2.0$ & $2.3\pm1.0$ & $3.4\pm1.0$ & $2.0\pm0.8$\\
        fs-vid2vid   &  $7.5\pm1.1$ & $1.8\pm0.5$ & $3.3\pm0.6$ & $2.2\pm0.5$\\ 
        StyleHEAT  &  $6.0\pm0.4$ & $1.4\pm0.2$ & $3.3\pm0.5$ & $\underline{1.5\pm0.2}$\\
        \textbf{Ours}  & $\mathbf{2.9\pm0.2}$ & $\mathbf{1.1\pm0.1}$ & $\mathbf{1.5\pm0.2}$ & $\mathbf{0.8\pm0.1}$\\
        \hline
    \end{tabular}
    }
    \caption{\textbf{Ablation on One-Shot Robustness.} We evaluate the robustness of each model in the task of same-identity re-enactment using 5 driving videos (500 samples/video) and 5 different source frames per driving video (25 source image-driving video combinations). Our approach yields the least mean and standard deviation proving its robustness. }
    \label{tab:robustness}
    \vspace{-0.1in}
\end{table}

\textbf{Pre-training Stage:} The entire Encoder, $E$, was trained during this stage on the CelebV-HQ dataset for 200K iterations.
The Feature Extraction Network, $F$, and the Identity Encoder, $E_{ID}$, were initialized with the pre-trained weights of e4e, and the training followed a progressive approach as proposed in \cite{tov2021designing}, where latent layers are progressively incorporated into the optimization at regular intervals. The Ranger optimizer (Rectified Adam \cite{liu2019variance} + Lookahead \cite{zhang2019lookahead}) was used to train the encoder, while Adam optimizer \cite{kingma2014adam} was used to train the latent discriminator. 

\textbf{Fine-tuning Stage:} Subsequent to the pre-training stage, the network was fine-tuned for 20K additional iterations on the HDTF training set with the addition of feature reconstruction losses to capture fine facial attributes. 
In this stage, the Feature Extraction Network is frozen and only the two latent-prediction heads: $E_{ID}$ and $E_{F}$ are fine-tuned at a reduced learning rate to avoid over-fitting. 
All hyperparameters remain unchanged, except $\lambda_{f}$ incorporating feature losses to the loss objective.


\textbf{Resources:}
 The network was trained on two RTX A6000 GPUs for approximately 100 hours.

\textbf{Inference Stage:} During inference, a driving video and a single source frame were obtained from the evaluation samples of the HDTF dataset. While the source frame and the driving video are of the same identity in same-identity re-enactment, the identities differ in the case of cross-identity re-enactment. The animated frames were generated as explained in \cref{eq:encode,eq:img_gen,eq:reenact,eq:edit}. Cyclic manifold adjustment (\cref{sub:cyc_manifold}) was used to improve realism and visual quality. Since a generative architecture is used, our model is capable of rendering re-enactment videos in real-time ($\sim30$fps).

\subsection{Baselines and Metrics}
\textbf{Baselines:}  We compare our results against a diverse range of state-of-the-art approaches that are based on: predictive keypoints (FOMM \cite{siarohin2019first}); 3D models (PIRenderer \cite{ren2021pirenderer}, StyleHEAT\cite{yin2022styleheat}), facial landmarks (fs-vid2vid\cite{wang2019few}); intermediate latents (LIA \cite{wang2022latent}); StyleGAN2-based (StyleHEAT); warping approaches (FOMM, fs-vid2vid, PIRender, StyleHEAT); and generative approaches (LIA).

\textbf{Metrics:} We extensively evaluate the results of the proposed framework against the baselines using 
(1) \textbf{reconstruction fidelity}: L1 norm pixel loss,  Peak Signal-to-Noise Ratio (PSNR),
(2) \textbf{Identity preservation:} Identity loss ($\mathcal{L}_{ID}$ - computed using \cite{deng2019arcface}), 
(3) \textbf{Perceptual quality:} LPIPS \cite{zhang2018perceptual},SSIM \cite{wang2004image}, FID \cite{heusel2017gans}, 
(4) \textbf{Spatio-temporal perceptual quality:} FVD \cite{unterthiner2018towards}, FVD\textsubscript{M} (FVD over the mouth), 
(5) \textbf{Temporal coherence in facial attributes:}  
$\rho_{\scaleto{AU}{3pt}}$, $\rho_{\scaleto{GZ}{3pt}}$, and  $\rho_{\scaleto{pose}{3pt}}$, the temporal correlation of Action Units (expressions), gaze, and pose respectively measured using \cite{openface2018}.

\subsection{Evaluation}

\textbf{Quantitative Analysis:} The performance of our approach in comparison to the baselines in the tasks of same-identity re-enactment and cross-identity re-enactment, are tabulated in \cref{tab:same_id_reenact,tab:cross_id_reenact} respectively. 
Our approach yields the best performance across all metrics except $\mathcal{L}_{ID}$ where it achieves comparable values to the best. A higher $\mathcal{L}_{ID}$ for FOMM is expected as it operates in the image space as opposed to our generative approach.
However, the performance of our model generating results at $1024^2$, in comparison to the baselines is not fully reflected through the metrics, as (1) most of the metrics ($\mathcal{L}_{ID}$, FID, FVD, \etc) are computed on downsampled images at a low-resolution which do not reflect the fine-grained details and (2) the lack of high-resolution ($\ge 1024^2$) data samples.  

\textbf{Qualitative Analysis:} The same-identity and cross-identity re-enactment examples  in \cref{fig:reenact} demonstrate that our approach yields much improved visual results compared to the baselines that are prone to visual artifacts, incorrect facial attributes, and lack of sharpness in scenarios where the source and driving frames have significant difference in pose and/or expression, subject is wearing eyeglasses, filling in unseen details (\eg teeth, ear). Our approach successfully handles these cases while producing more realistic re-enactment  at high-resolution ($1024^2$). 
The versatility of our model is more pronounced  in the cross-identity examples in \cref{fig:reenact} \textit{(Bottom)}, where facial deformations of the driving frames are accurately mimicked in the re-enacted frame while preserving the source identity. 
Realistic re-enactments with latent-based attribute edits such as beard, age, make-up, \etc applied to the source image (see \cite{harkonen2020ganspace,shen2020interface}) could be generated as shown in \cref{fig:teaser}. This is enabled by the design choice of $\mathcal{W}_{ID}$ being in the well-editable $W+$ space.

Based on the visual results,  artifacts are most prominent in warping based approaches (FOMM, fs-vid2vid, PIRenderer, StyleHEAT). Further, all approaches except StyleHEAT are only capable of low-resolution generation. While StyleHEAT uses StyleGAN2 to generate re-enactment videos at $1024^2$, the use of a 3D model to capture the facial attributes and warping of the intermediate feature spaces inhibits its performance. In contrast, our model exploits implicit priors of the ore-defined latent spaces of StyleGAN2 to achieve state-of-the-art performance. Please refer to the video examples on the project page. 

\begin{figure}
    \centering
    \includegraphics[width=\linewidth]{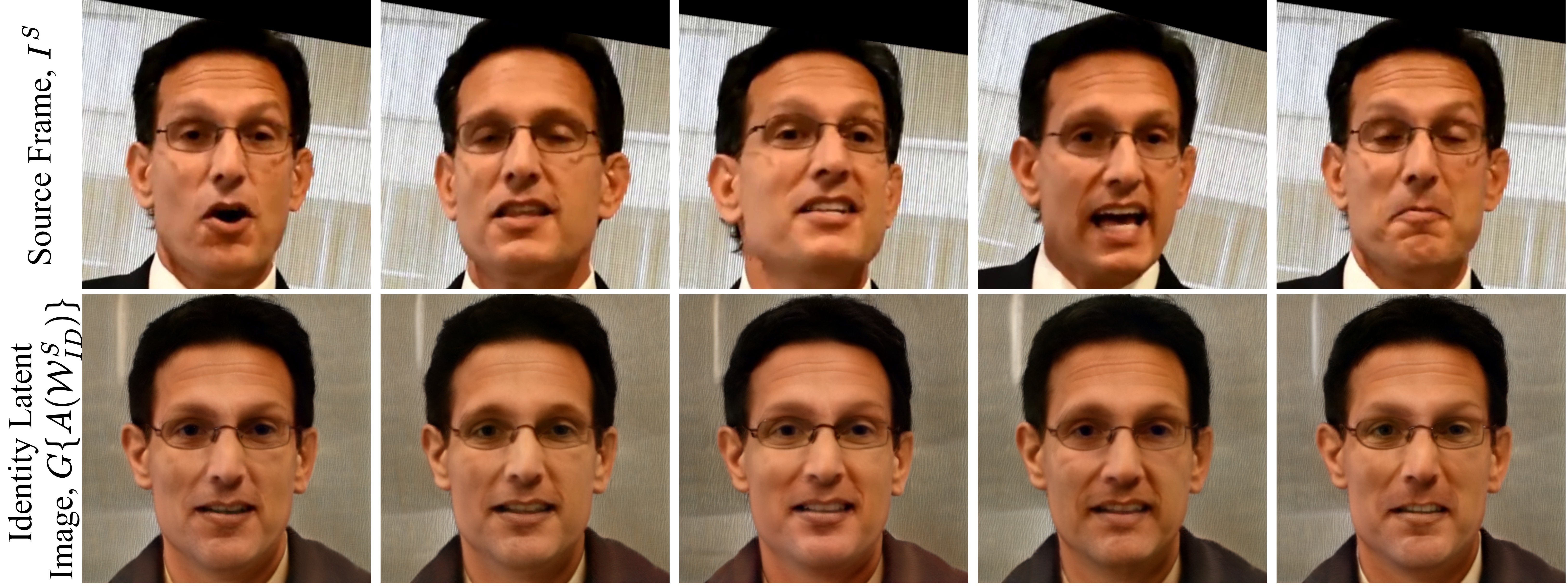}
    \caption{\textbf{Identity Latent Images.} Each column consists of the source frame \textit{(Top)} and the corresponding Identity latent image \textit{(Bottom)}. Even though the source images significantly differ in head-pose, expression, and eyes, the identity latent captured is nearly identical (with neutral pose and expression) which contributes towards the robustness of our approach.}  
    \label{fig:id_latent}
    \vspace{-0.1in}
\end{figure}

\subsection{Ablation Study}
\textbf{Identity Regularization:} We evaluate the impact of using the identity-loss based regularization terms (\cref{eq:l_id_s,eq:l_wid}), which are in place to minimize the identity leakage into the facial deformation latent, $\mathcal{S}_F$. The results are in   \cref{tab:same_id_reenact,tab:cross_id_reenact} (\textit{Bottom}) as \textit{Ours w/o ID reg}.   Considerable quantitative improvement is seen in cross-identity re-enactment with the inclusion of identity regularization.

\textbf{Hybrid Latent Spaces:} We propose a hybrid latent approach, where the Identity latent, $\mathcal{W}_{ID}$, and the Facial deformation latent, $\mathcal{S}_F$, reside in the domains of $W+$ and $SS$ respectively to make use of the editability and reconstruction of the $W+$ space and the disentanglement of the $SS$. We validate the use of a hybrid approach against \textit{Ours w/o Hybrid}, where both the Identity and Facial deformation latents reside in the $W+$ space. Based on the results at the bottom of \cref{tab:same_id_reenact,tab:cross_id_reenact}, it could be observed that the scores deteriorate compared to the Hybrid approach, most likely due to the entanglements within the $W+$ space.

\textbf{Cyclic Manifold Adjustment (CMA):}
We evaluate the performance improvement of using Cyclic Manifold Adjustment in comparison to (1) our framework without CMA (\cref{tab:cross_id_reenact}: \textit{Ours w/o CMA}) and (2) replacing CMA with PTI\cite{roich2021pivotal} (\cref{tab:cross_id_reenact}: \textit{Ours -- CMA + PTI}). While using CMA achieves the best results, it could be observed that the use of PTI deteriorates the scores compared to \textit{Ours w/o CMA}. Using PTI on the source does not guarantee the local manifold around the source to accommodate the facial deformations of the driving sequence leading to distorted/ suboptimal results. In contrast, our approach  improves the identity reconstruction of the out-of-domain source and also enables seamless transfer of facial deformations of the driving video to the source, as we tweak the generator such that the driving facial deformation edits reside within the local manifold centered around the source identity latent.

\textbf{One-shot Robustness:} The robustness to different head-poses and facial attributes of the source frame is evaluated. The performance of same-identity one-shot re-enactment of 5 driving videos with 5 different source frames per driving video (25 source image-driving video combinations) are evaluated (see  \cref{tab:robustness}). 
We observe that our approach is the most robust with the lowest mean and standard deviation. 

We attribute the robustness to the Identity and Facial deformation decomposition we employ. As shown in \cref{fig:id_latent} the network implicitly learns an identity latent that has a neutral pose and expression irrespective of the head-pose and facial attributes of the source. While most algorithms attempt to directly capture the difference between the source and driving frames, in contrast, we encode the facial deformations relative to an implicitly learned neutral pose and identity. While LIA follows a similar approach of anchoring the facial deformations to an implicitly learnt reference, their method requires the source and the first frame of the driving sequence to have similar pose and is limited to $256^2$.

\subsection{Limitations}
Since we base our model on StyleGAN2, we inherit its limitations of texture sticking and alignment requirements. Further, handling occlusions and reconstruction of changing backgrounds are challenging since the StyleGAN generator is pre-trained for faces.
While our model could be adapted to StyleGAN3\cite{karras2021aliasfree} to mitigate the issue of texture sticking, the use of StyleGAN2 is preferred due to its latent space being more structured and expressive \cite{alaluf2022third}.

\subsection{Negative Societal Impact}\label{sec:neg_societal}

The negative societal impact of our model is similar to that of other DeepFake algorithms. \ie, the impressive performance of the proposed model in one-shot re-enactment entails the risk of the model being used with malicious intent. However, the advancements in re-enactment research create social awareness in the community of such methods and also pave the path for research on DeepFake detectors \cite{marra2018detection, gragnaniello2021gan,wang2022gan,zhang2022exposing}. Further, the lack of high resolution datasets prevents our model from reaching its full potential. Moreover, the aforementioned limitations of StyleGAN such as texture sticking and limitations in background reconstruction would provide cues for DeepFake detectors.

\section{Conclusion}
We propose an end-to-end unified framework to facilitate high-fidelity one-shot facial video re-enactment at $1024^2$, exploiting the implicit priors in StyleGAN2 latent spaces. 
The framework reveals the full potential of StyleGAN2 for face edits, specifically identity, attributes, and facial deformations in videos.
The model is centered around hybrid latent spaces to encode the identity and facial deformations that exploit the editability, reconstruction, and disentanglement properties of each latent space, thus achieving state-of-the-art results both quantitatively and qualitatively. 
Moreover, the proposed model is robust to diverse head-poses and expressions of the source frame.
Further, the model facilitates generation of re-enactment videos with latent-based edits (beard, age, make-up, \etc) proposed by previous research.

{\small
\bibliographystyle{ieee_fullname}
\bibliography{egbib}
}

\end{document}